\documentclass[sigconf]{acmart}

\AtBeginDocument{%
  \providecommand\BibTeX{{%
    \normalfont B\kern-0.5em{\scshape i\kern-0.25em b}\kern-0.8em\TeX}}}

\usepackage{comment}

\setcopyright{rightsretained}

\begin{document}
\fancyhead{}
\title{Continual Learning of Visual Concepts for Robots through Limited Supervision}


\author{Ali Ayub}
\email{aja5755@psu.edu}
%\author{Alan R. Wagner}
%\email{alan.r.wagner@psu.edu}
\affiliation{%
  \institution{The Pennsylvania State University}
  %\streetaddress{P.O. Box 1212}
  \city{State College}
  \state{PA}
  \country{USA}
}
\author{Alan R. Wagner}
\email{alan.r.wagner@psu.edu}
\affiliation{%
  \institution{The Pennsylvania State University}
  %\streetaddress{P.O. Box 1212}
  \city{State College}
  \state{PA}
  \country{USA}
}

%\renewcommand{\shortauthors}{Ayub et al.}


\begin{abstract}
  For many real-world robotics applications, robots need to continually adapt and learn new concepts. Further, robots need to learn through limited data because of scarcity of labeled data in the real-world environments. To this end, my research focuses on developing robots that continually learn in dynamic unseen environments/scenarios, learn from limited human supervision, remember previously learned knowledge and use that knowledge to learn new concepts. I develop machine learning models that not only produce State-of-the-results on benchmark datasets but also allow robots to learn new objects and scenes in unconstrained environments which lead to a variety of novel robotics applications. 
  
\end{abstract}

\copyrightyear{2021}
\acmYear{2021}
%\setcopyright{rightsretained}
\acmConference[HRI '21 Companion]{Companion of the 2021 ACM/IEEE International Conference on Human-Robot Interaction}{March 8--11, 2021}{Boulder, CO, USA}
\acmBooktitle{Companion of the 2021 ACM/IEEE International Conference on Human-Robot Interaction (HRI '21 Companion), March 8--11, 2021, Boulder, CO, USA}
\acmDOI{10.1145/3434074.3446357}
\acmISBN{978-1-4503-8290-8/21/03}

\maketitle

\section{Introduction}
\label{sec:introduction}
Continual adaptation and learning through limited data is the hallmark of human intelligence. Humans continue to learn new concepts over their lifetime without the need to relearn most previous concepts. With robots becoming an integral part of our society, they must also continue to learn over their lifetime to adapt to the ever-changing environments. Further, in real-world applications, robots do not have access to a large amount of labeled data since it is impractical for human users to provide hundreds of examples to the robot. Thus, robots must learn using a small amount of data through limited human supervision. The long-term goal of my research is to develop autonomous robots for everyday environments where they can learn over their lifetime and use the learned knowledge to assist humans in their daily lives.%My work focuses on answering two main questions: How can robots 1) learn visual concepts (objects and scenes) continually 2) learn from limited human supervision. %In this work, I focus on developing machine learning models that allow robots to 1) continually learn visual concepts (objects and scenes) in dynamic unseen environments 2) learn from limited human supervision. 

Creating robots that continually learn is a challenging problem. Deep learning is widely used to address many robot learning tasks, yet deep learning suffers from a phenomenon called \textit{catastrophic forgetting} when learning continually. Catastrophic forgetting occurs when continually training a model (neural network) to recognize new classes, the model forgets the previously learned classes and the overall classification accuracy decreases. One way to address this problem is by storing the complete data of the previously learned classes. However, storing data of the previous classes requires a huge memory when learning new classes continually. Robots, on the other hand, have limited on-board memory available, hence they cannot keep storing high-dimensional images of previous classes. In real-world scenarios, labelling a large amount of data is costly in terms of time and effort. Hence, robots have to learn from a small number of interactions with likely impatient human users. Deep learning systems, however, require a large amount of labelled data for learning.

In order to tackle these challenges, my work develops machine learning and computer vision techniques that are inspired by concept learning models from cognitive science.
My work is informed by higher level concept learning in children (and all humans) related to curiosity-driven, intrinsically motivated, continual learning of visual concepts (objects and scenes). %My research focuses on the creation of cognitively inspired frameworks, that allow robots to learn new concepts (objects, scenes) continually through limited data and human supervision. 

%Pose the problem just like the curiosity driven paper. Then talk about the application a little bit. And then move toward the solution.Background wise: just give some background on continual learn-ing and maybe few-shot learning and some curiosity Approach: Want to give a broader aspect of the approach. Tell them your main approach and how do you go towards solving such a problem. So essentially you also want to say how it is applicable on the robots.

%this should explain the key questions tackled by the paper. Introduce the problem. And then talk about what questions you plan to address in this paper. Essentially, the main goal of this work is to create continual learning techniques that allow the robot to learn visual concepts through limited human supervision in unconstrained environments.  
\section{Related Work}
\label{sec:related_work}

Recent continual learning techniques use deep neural networks and rely on storing a fraction of old class data when learning a set of new classes~\cite{Rebuffi_2017_CVPR,Castro_2018_ECCV}. To avoid storage of real samples, some approaches use generative-memory and regenerate samples of old classes using GANs or autoencoders \cite{Ayub_ICML_20,Ayub_iclr20,Ostapenko_2019_CVPR,Ayub_iclr20}, however the performance of these approaches is generally inferior to approaches that store real images. One major issue with all these prior continual learning approaches is that they require a large amount of training data. Hence using these methods for continual learning from limited data results in poor accuracy. 

Curiosity-driven learning has been explored for robotics applications in the past to learn from limited data and supervision. %Among early works, \cite{schmidhuber91}, \cite{Ugur07} and \cite{oudeyer07} present curiosity-driven reinforcement learning approaches for robot controllers. These approaches rely on classical techniques and are not compatible with deep network architectures. 
In recent years, some deep reinforcement learning approaches have been proposed that use a curiosity-driven reward function \cite{burda18,haber18} to train neural networks. For object learning, many researchers have presented active learning techniques using uncertainty sampling \cite{Beluch_2018_CVPR,Gal17,Siddiqui_2020_CVPR,Yoo_2019_CVPR,Shen_2019_ICCV}. All of these approaches train deep networks using specific loss terms such that the network can predict the most uncertain samples. Although these approaches produced good results on small, simple image datasets like MNIST \cite{Lechun98}, they were not tested on a real robot. 

One of the main limitations of prior curiosity-driven and active learning approaches is that they are designed for a batch learning setting and will thus suffer from catastrophic forgetting when attempting to learn continually. %Note that even though deep networks are trained in a batch setting to predict new unknown classes in an increment, these models lose their ability to recognize the unknown data in subsequent increments because the network begins to assign the unknown classes to the newly learned classes that were unknown in the previous increments. 
In contrast, we present a novel approach that not only allows a robot to learn from visual data continually but also allows it to assign curiosity scores to unlabeled objects in a self-supervised manner.  

%Related work should talk about some continual learning approaches and their issues. 

%Then the problem of continual learning in robots and its issues. 

%Finally moving towards some curiosity based or active learning, few-shot learning approaches but they do not learn continually.
\section{Methodology}
\label{sec:methodology}
In this work, I consider a general continual learning setup for learning visual categories (object or scene classes). In each new increment $t$, the robot gets a small set of labeled samples $S_t = \{(x_i^t,y_i^t)\}_{i=1}^{n^t}$, where $x_i \in \mathcal{X}$ are the visual samples (images) and $y_i^t$ are their ground truth labels. The samples in an increment can belong to the earlier learned classes or completely new classes. Further, the robot has limited storage capacity, thus it cannot store the high dimensional images of the previously learned categories. 

To learn new objects or scenes, the robot first acquires new image data autonomously using its own cameras.The category labels for the images are provided by the human in a textual format. I then use a neural network pre-trained on a large dataset (e.g. ImageNet \cite{Russakovsky15}) to extract feature vectors for the images. Then, I apply a novel cognitively-inspired clustering approach (called \textit{Agg-Var} clustering) on the feature vectors of the images to learn centroids and covariance matrices for the visual categories. In \textit{Agg-Var} clustering, the model finds the Euclidean distance of a new $i$th feature vector $x_i^y$ of a class $y$ to the previously learned centroids of the class. If the distance is below a pre-defined distance threshold $D$ (hyperparameter), the model performs \textit{memory integration} \cite{Mack18} by updating the closest centroid and the corresponding covariance matrix using the new feature vector. If the distance is above the distance threshold $D$, the model performs \textit{pattern separation} \cite{Mack18} by creating a new centroid initialized with $x_i^y$ and a new covariance matrix initialized with a zero matrix. 
%In this clustering technique, the model either initializes a new centroid/covariance matrix pair for a category using new image data, if the category is previously unknown, or it updates the closest centroid/covariance matrix pair using the new image data for the class. Euclidean distance between centroids and feature vectors is used to find the closest centroids. 
In this way, the model gets a set of centroids and covariance matrices for all the classes separately. Note that even a small number of images per class are enough to learn the centroids/covariance matrix representation for the class, hence my model can be used to learn from limited labeled data. 

For classification of test images, I use pseudorehearsal technique \cite{Robins95} in which I use the centroids and covariance matrices of the old classes as parameters of Gaussian distributions. I then sample these Gaussian distributions to generate pseudo-exemplars for the old classes. A shallow neural network classifier with a single linear layer is then trained using the pseudo-exemplars and the feature vectors of the images in the current increment. In this way, the model mitigates catastrophic forgetting. 

%What is the methodology?

%Actually, talk about how you represent the problem and then use the cognitively inspired technique. Mostly the curiosity approach. 

%Well, we come up with a new model inspired by the human cognitive system to allow the model to learn continually, with a few samples, using limited memory.

%The model also allows the robot 

\section{Past, Current and Future Work}
\label{sec:work}
Towards the goal of creating continually learning robots, the first project in my PhD was focused on the few-shot incremental learning problem (FSIL), in which the robot learns continually from a small number of object examples provided by a human. %One limitation in this setup was that the robot was provided with the objects belonging to a class in a single increment. 
I developed a novel approach termed Centroid-Based Concept Learning (CBCL) %the preliminary version of the %CBCL-PR approach (termed CBCL) 
to tackle this problem \cite{Ayub_2020_CVPR_Workshops}. CBCL’s classification accuracy was significantly higher than the State-of-the-art (SOTA) incremental learning approaches on benchmark datasets (Table \ref{tab:results}). I then applied CBCL on a real robot for a cleaning application, in which the robot learns household objects from a few visual examples provided by a human and organizes related objects from a clutter of objects. This research demonstrated that my method could be capable of dynamically learning task or situation specific objects \cite{Ayub_IROS_20}. I also showed that CBCL is a general approach and can be applied for other tasks, such as RGB-D indoor scene classification \cite{Ayub_2020_BMVC}.

In real-world environments, robots must learn from streaming data that lacks well-defined task boundaries (online learning). The lack of task boundaries and unknown number of categories makes this problem harder than FSIL. I developed an updated version of CBCL, termed Centroid-Based Concept Learning with Pseudorehearsal (CBCL-PR) for online learning. CBCL-PR significantly outperformed SOTA approaches on a benchmark dataset in terms of detecting known and unknown scene categories. I then applied CBCL-PR on the Pepper robot in which the robot wandered in unconstrained real-world environments to learn new scene categories and detect previously unknown scene categories \cite{Ayub_corl20}.

\begin{table}
\centering
\small
\begin{tabular}{ |p{1.7cm}|p{1.0cm}|p{1.0cm}|p{1.0cm}|p{1.0cm}|}
     \hline
    \textbf{Methods} & iCaRL & EEIL & BiC & \textbf{CBCL} \\
     \hline
    \textbf{Accuracy (\%)} & 63.75 & 64.02 & 64.84 & \textbf{69.85}\\
 \hline
 \end{tabular}
 \caption{Comparison of CBCL with iCaRL \cite{Rebuffi_2017_CVPR}, EEIL \cite{Castro_2018_ECCV} and BiC \cite{Wu_2019_CVPR} for class-incremental learning with 10 classes per increment on the CIFAR-100 dataset \cite{Krizhevsky09}.}% The accuracies represent average results over 10 increments.}
 \label{tab:results}
 \end{table}

In follow-up research, I developed a system that used CBCL-PR for online learning of scenes/contexts and Dempster Schaeffer theory to represent and learn appropriate norms related to different scenes in terms of conditional probabilities. My work was the first of its kind to examine online learning of norms for social robots. I tested this approach on Pepper in which the robot wandered around at different scenes and learned norms through simple Q/A sessions with a human.% the robot successfully learned the appropriate norms at each scene category in an online manner.
This research demonstrated that my approach may allow robots to learn different scene categories and use the recognition of these scenes to moderate their behavior and decision-making %For example, allowing a robot to recognize that its current location is in a library and that it should therefore reduce the volume of its speech 
\cite{Ayub_ICSR20}.

%For robots to learn in real-world environments, they usually have access to a large amount of unlabeled data, and they need to ask humans for supervision. Since it is difficult for humans to answer an enormous number of questions, robots must continually learn in a curiosity driven manner, i.e. they should prioritize and only ask about the concepts they are most uncertain about from humans.
I am currently working on the curiosity-driven active online learning (CDAOL) problem, in which the robot has a large amount of unlabeled objects available in an environment and it must choose the most informative samples to be labeled. %I am developing a novel technique to train a convolutional neural network (CNN) using a loss function inspired by CBCL that generates centroids for the learned classes. %To avoid the centroids of old classes to drift when training the CNN on new classes, I stored a single prototype image per centroid and reused it during the training of new classes. 
%For assigning curiosity scores to new objects in a self-supervised manner, I use the distance of the unlabeled objects from the previous centroids. %two intuitions: 1) the robot should be more curious about objects that are farther apart from the previous centroids, 2) the robot should be more curious about such objects whose different views are assigned different labels by the CNN. 
%I tested my approach on benchmark datasets and it significantly outperformed the SOTA continual learning approaches in terms of classification accuracy. 
I am developing a novel approach to assign curiosity scores to new unlabeled objects in a self-supervised manner using the distance of the new objects from the previously learned centroids. Preliminary experiments show that my approach can learn the most informative objects quickly without forgetting the previously learned objects which results in a dramatic increase in accuracy over the other approaches, especially in the earlier increments \cite{Ayub_RoMan_20}. 

For future work, I plan to apply the above-mentioned approach on a real robot. However, real-world robots have access to clutters of objects rather than single object images. Second, capturing multiple views of individual objects in unconstrained environments through robot's own cameras without human assistance is challenging. To deal with this, I plan to develop a complete system to allow a robot to capture images of cluttered objects, localize all the objects in the clutter, get labels for the most informative objects, use a manipulator module to move its hands around the labeled objects to get different views of the objects and finally train the CNN using the images of the new objects. I plan to test this system on a real robot in a lab environment with clutter of objects present at various locations with different backgrounds. The experiment will be performed over the course of one month at different times of the day in which the robot will wander around in the environment and learn about the objects it is curious about by asking a human teacher. This experiment is the first of its kind, that will demonstrate a true lifelong learning robot that learns a large number of objects (240 objects) in an unconstrained environment over a long period of time through limited human supervision.

%Talk about what you have done. FSIL with robotics application. Online learning of scenes with updated CBCL-PR, also learn context-specific norms. And finally the next project is to use the representations to learn curiously. 

  \bibliographystyle{ACM-Reference-Format}
  \bibliography{main}

\end{document}